\documentclass{article} 
\usepackage[preprint]{colm2026_conference}

\usepackage{microtype}
\usepackage{hyperref}
\usepackage{url}
\usepackage{booktabs}
\usepackage{amsmath}

\usepackage{lineno}
\usepackage{graphicx}
\usepackage{multirow}
\usepackage{subcaption}
\usepackage{adjustbox}
\usepackage[normalem]{ulem}

\usepackage[table]{xcolor}
\usepackage{algorithm}
\usepackage{algpseudocode}

\definecolor{lightblue}{RGB}{220,235,250}
\definecolor{darkblue}{rgb}{0, 0, 0.5}
\hypersetup{colorlinks=true, citecolor=darkblue, linkcolor=darkblue, urlcolor=darkblue}

\makeatletter
\AtBeginDocument{
\renewcommand{\sectionautorefname}{\S\@gobble}

\renewcommand{\subsectionautorefname}{\S\@gobble} 
\renewcommand{\subsubsectionautorefname}{\S\@gobble}
\renewcommand{\appendixautorefname}{Appendix \S\@gobble}

}
\makeatother

\title{From the Inside Out: Progressive Distribution Refinement for Confidence Calibration}

\author{%
  Xizhong Yang$^1$\thanks{Work done during an internship at Kuaishou Technology.}\quad
  Yinan Xia$^2$\quad
  \setcounter{footnote}{1}
  Huiming Wang$^3$\thanks{Corresponding author: Huiming Wang, Mofei Song.}\quad 
  $\text{Mofei Song}^{1\dag}$ \\[6pt]
  $^1$Southeast University, \quad $^2$Kuaishou Technology, \quad $^3$SUTD\\
  $^\dag$\texttt{huiming\_wang@mymail.sutd.edu.sg}, \quad $^\dag$\texttt{songmf@seu.edu.cn}
}

\begin{document}

\ifcolmsubmission
\linenumbers
\fi

\maketitle

\begin{abstract}
    Leveraging the model's internal information as the self-reward signal in Reinforcement Learning (RL) has received extensive attention due to its label-free nature. While prior works have made significant progress in applying the Test-Time Scaling (TTS) strategies to RL, the discrepancy in internal information between test and training remains inadequately addressed. Moreover, Test-Time Training based on voting-based TTS strategies often suffers from reward hacking problems. To address these issues, we propose \textit{DistriTTRL}, which leverages the distribution prior of the model's confidence during RL to progressively optimize the reward signal, rather than relying solely on single-query rollouts. Additionally, we mitigate the phenomenon of consistent reward hacking caused by the voting-based TTS strategies through diversity-targeted penalties. Benefiting from this training mechanism where model capability and self-reward signals complement each other, and the mitigation of reward hacking, \textit{DistriTTRL} has achieved significant performance improvements across multiple models and benchmarks. Code available at  \href{https://github.com/yxizhong/DistriTTRL}{https://github.com/yxizhong/DistriTTRL}.
\end{abstract}
\section{Introduction}
    The development of techniques such as Chain-of-Thought~\citep{CoT} and Reinforcement Learning (RL) has significantly contributed to the performance improvement of Large Reasoning Models (LRMs)~\citep{DeepSeek-r1, OpenAI-o1}. Inspired by Reinforcement Learning with Verifiable Rewards (RLVR)~\citep{DeepSeek-r1}, an increasing number of works choose to scale up computational and data resources for training in verifiable scenarios, such as mathematics and coding. Nevertheless, unpredictable test environments largely limit the practical performance of models, a phenomenon that is particularly pronounced in the complex and variable application scenarios of LRMs. To address the insufficient generalization caused by LRMs' heavy reliance on human supervision, many current works turn to Test-Time Training (TTT)~\citep{TTT-ICL,TTT-Titans} on unlabeled test-time data to achieve model adaptation~\citep{TTT-Survey}.

    Current works that improve LRMs' reasoning capabilities through TTT primarily focus on how to provide valuable immediate rewards to models under label-free conditions. For instance, TTRL~\citep{TTRL} employs a voting-based Test-Time Scaling (TTS) strategy, Majority Voting~\citep{Self-Consistency}, which selects the most likely correct answer based on commonalities from multiple sampled results of a single question as the pseudo-label for that question at training stage, and further utilizes it to compute rewards. SCOPE~\citep{SCOPE} further uses the model's internal confidence to assist the voting process, improving TTT effectiveness by mitigating confirmation bias and sparse rewards. However, applying TTS strategies~\citep{BoN,MoB} designed for fixed-parameter models at test-time to the training process may reduce their effectiveness.

    We believe one reason is that limited computational resources constrain the number of samples per query. Although TTRL mitigates this through downsampling, it still hinders voting accuracy. More importantly, dynamically updated parameters during training cause continuous changes in the model's internal information (such as confidence and sampling distribution~\citep{Nips6666-Does}), preventing rollouts sampled at different steps from being mutually utilized. Additionally, updatable parameters allow the model to learn inherent hacking information in TTS strategies during training, such as consistency.

    \begin{figure}[t]
        \begin{center}
        \centerline{\includegraphics[width=0.98\columnwidth]{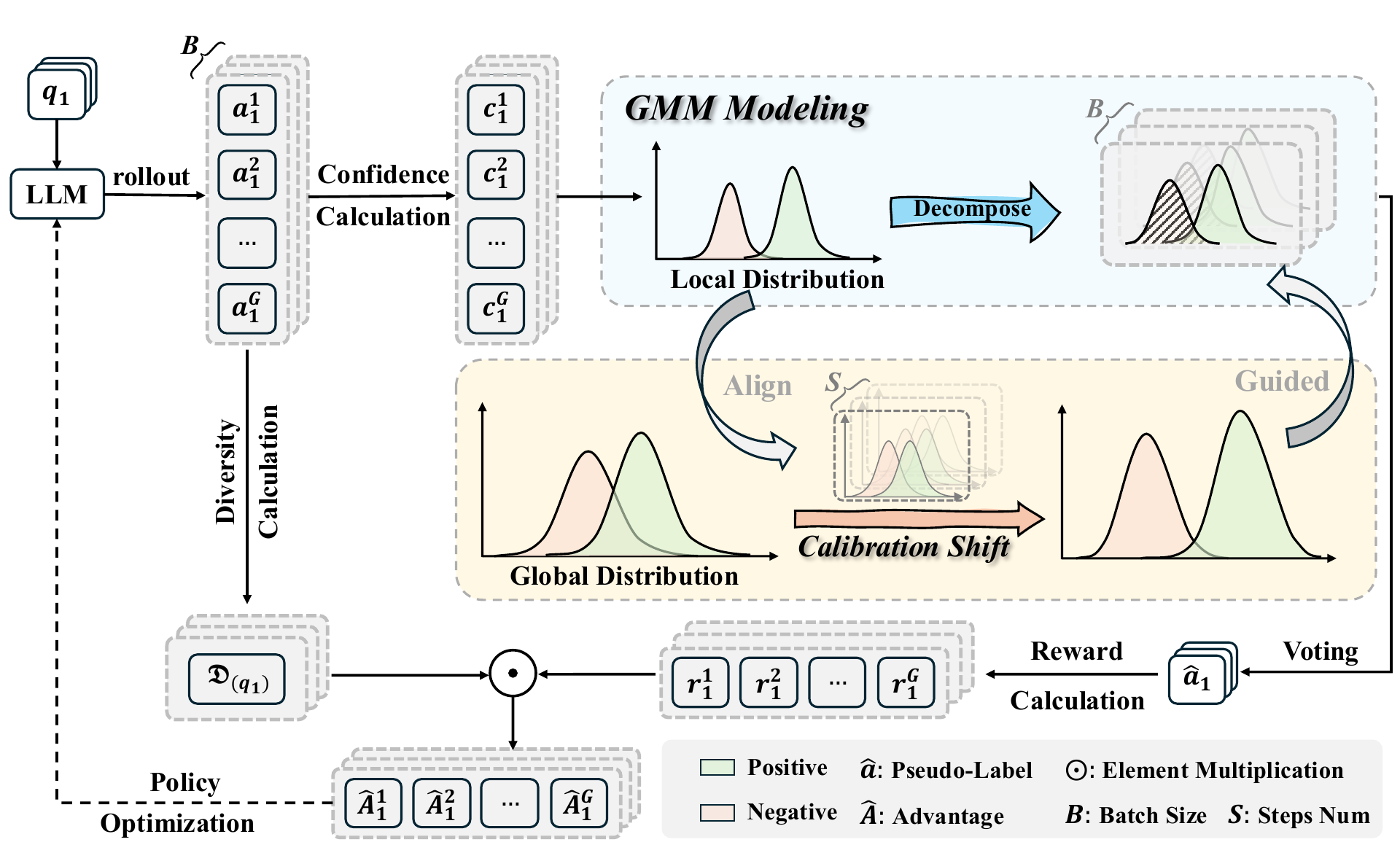}}
        \caption{
            Overview of \textit{DistriTTRL}. During the pseudo-label estimation process at each step, we first calibrate the Global Confidence Distribution from previous steps using the Local Confidence Distribution of all queries in the current step, then guide the distribution of specific queries in the current step with the calibrated distribution prior. Additionally, we computes diversity from each query's rollouts and adjusts the advantage accordingly.
        }
        \vskip -0.4in
        \label{fig:main}
        \end{center}
    \end{figure}

    Motivated by these, we propose \textit{DistriTTRL}, which improves the accuracy of RL rewards by progressively constructing distributions of rollouts information at different steps, and designs diversity-targeted penalties to mitigate the consistency reward hacking problem introduced by TTS strategies in TTT. Specifically, following DistriVoting~\citep{DistriVoting}, we weight answers using the model's internal confidence during the voting process and leverage distribution priors for confidence calibration. Building upon this, to improve the stability of confidence distribution prior information by increasing the number of samples, we dynamically record all rollouts at each step, progressively constructing an increasingly refined confidence distribution. Furthermore, since parameter updates between different steps cause distribution shifts, we perform shift correction on the confidence of each previous step based on the current step's distribution during construction. After obtaining pseudo-labels for each query through voting, we compute rewards similarly to other RLVR works. Differently, we subsequently evaluate the diversity of each query through the number of predicted answer classes in its rollouts and normalize this diversity metric within a training batch, which is then used as a penalty weight for the advantage of each query in the batch.

    In the experimental phase, we validate our approach using multiple models including Qwen2.5/Qwen3-Series~\citep{Qwen3}, and DeepSeek-Series~\citep{DeepSeek-r1} across different benchmarks including AIME~\citep{AIME}, AMC~\citep{AMC}, MATH-500~\citep{MATH-500}, and GPQA-D~\citep{GPQA-D}. The results consistently demonstrate the effectiveness of DistriTTRL. Moreover, by analyzing the majority ratio (i.e., the frequency proportion of the voted answer among all rollout answers) of voting results for each query during training, we can confirm that our designed diversity-targeted penalty effectively mitigates consistency reward hacking.
\section{Related Work}
\subsection{Test-time Training}
    Test-Time Training~\citep{liu2021ttt++} has emerged as a focus of recent research in the context of generative tasks~\citep{shocher2018zero,bau2020semantic}. By leveraging self-supervised learning (SSL)~\citep{liu2021self}, this emerging paradigm enables models to cope with distributional shifts without requiring additional labeled data. Recent work~\citep{TTRL,zhang2025right} on large language models~\citep{DeepSeek-r1,Qwen3} has begun to leverage test-time training during the reinforcement learning stage, enabling models to learn from unlabeled data without relying on costly manual annotations. However, they rely on voting-based TTS strategies for test-time training, which often leads to reward hacking due to misaligned incentives in the reinforcement learning process.

\subsection{Intrinsic Information of LLMs}
    Intrinsic information denotes the model’s internal confidence signals derived from next-token distribution statistics, has recently been leveraged to assess the quality of reasoning traces~\citep{geng2024survey,fadeeva2024fact,tao2024trust}. Owing to its label-free nature, intrinsic information has been applied in training to extend RLVR methods~\citep{EndoRM, TTRL} without relying on costly manually labeled data, and during inference to optimize multi-candidate responses through serial, parallel, and tree-based search. Several methods have been proposed to quantify intrinsic information: Average Log-Probability~\citep{DeepConf} and Perplexity~\citep{horgan1995complexity} capture sentence-level confidence, while Entropy~\citep{renyi1961measures} and Self-Certainty~\citep{Self-Certainty} measure distribution-level confidence across multiple candidates.
\section{Preliminaries}

\subsection{Confidence of Trajectory}
    Prior work shows that LLM's token distributions $\mathbf{P}_i(j)$ can reveal uncertainty and trajectory quality. Following DeepConf~\citep{DeepConf}, we compute confidence via token negative log-probabilities to assess quality during and after inference. For a single trajectory containing $N$ tokens output by the model, we define the trajectory confidence as:
    \begin{equation}
    \label{eq:traj_conf}
        C_{\text{traj}} = -\frac{1}{N_G \times k}\sum_{i \in G}\sum_{j=1}^k \log \mathbf{P}_i(j),
    \end{equation}
    
    where $N_G, k \in \mathbb{N}^+$ and $\mathbf{P} \in \mathbb{R}^{N \times N_v}$.
    Here, $G$ represents the subset of tokens among the $N$ generated tokens that are used to compute the final trajectory-level confidence, which typically corresponds to the last tail step containing the answer. $N_G$ denotes the number of tokens in $G$, and $k$ denotes the number of top-$k$ probabilities from the token logits that participate in computing token-level confidence.

\subsection{Distribution of Confidence}
\label{sec:preliminary_conf_distribution}
    For each query's $G$ rollouts, we compute trajectory-level confidence scores $\mathcal{C}$ via \autoref{eq:traj_conf}. Inspired by previous work on confidence distribution~\citep{Self-Certainty}, we further model it as Gaussian Mixture Models (GMM)
    with two components, where the positive distribution $\mathcal{N}(\mu_{\text{pos}}, \sigma_{\text{pos}}^2)$ with a higher mean $\mu_{\text{pos}}$, and the negative distribution $\mathcal{N}(\mu_{\text{neg}}, \sigma_{\text{neg}}^2)$ with a lower mean $\mu_{\text{neg}}$:
    \begin{equation}
        X_{\text{pos}} \sim \mathcal{N}(\mu_{\text{pos}}, \sigma_{\text{pos}}^2), \quad
        X_{\text{neg}} \sim \mathcal{N}(\mu_{\text{neg}}, \sigma_{\text{neg}}^2).
    \end{equation}

\subsection{Distribution Shift During the Training Process}
\begin{figure}[ht]
        \begin{center}
        \centerline{\includegraphics[width=0.65\columnwidth]{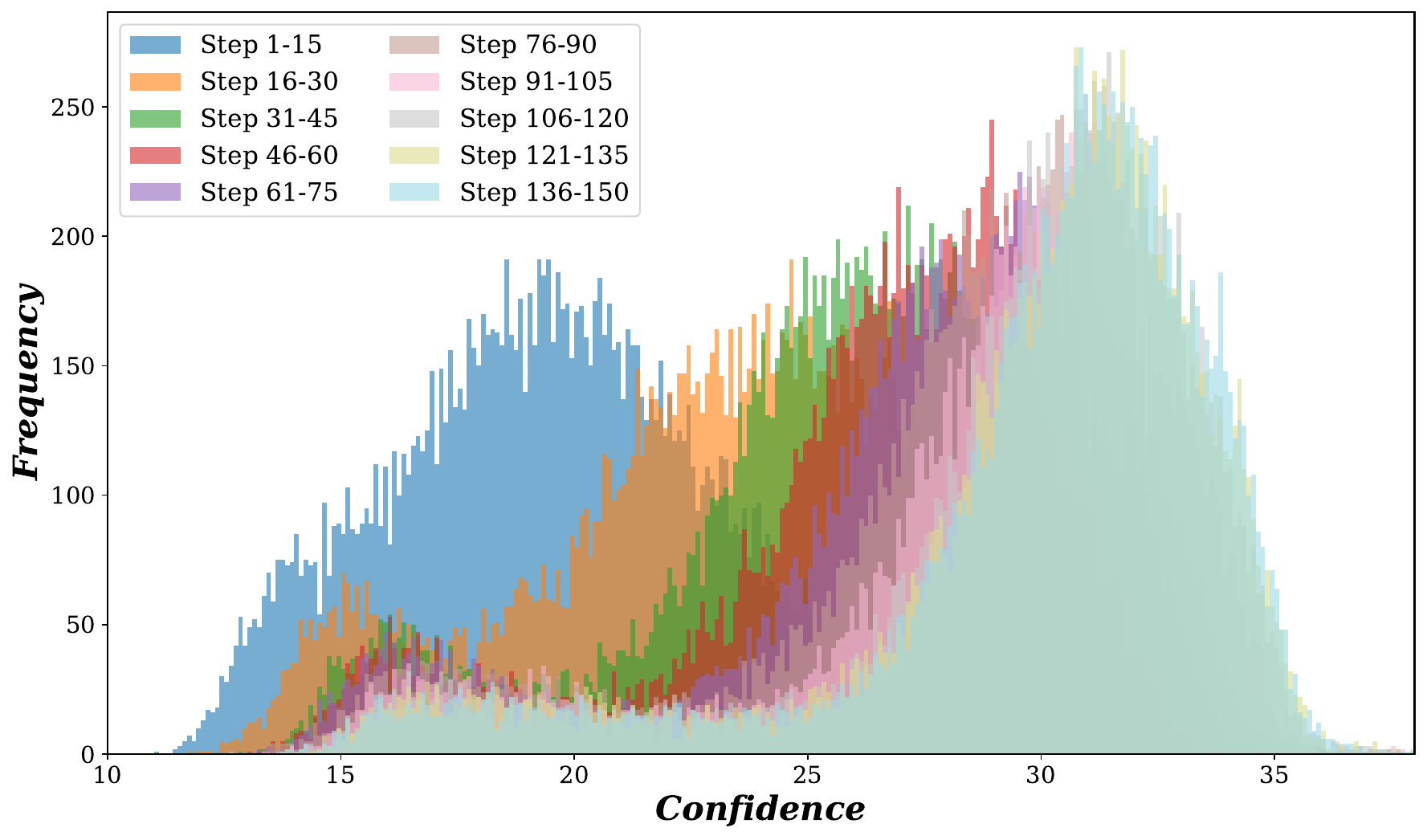}}
        \vskip -0.1in
        \caption{
            Confidence distribution shift across training steps. Trained on AMC using Qwen3-8B with 32 samples per question. Each distribution aggregates 15 consecutive steps.
        }
        \vskip -0.4in
        \label{fig:distribution_shift}
        \end{center}
    \end{figure}

    Unlike the Test-Time Scaling task where model parameters are fixed, the confidence distribution obtained at different stages of model training varies with model parameter updates, and this variation significantly affects the auxiliary calibration role of the global distribution prior on the current step. To analyze this variation, we record and visualize the model's confidence at different steps, as shown in \autoref{fig:distribution_shift}. As training progresses, \textbf{\textit{the confidence distribution gradually shifts rightward, with the degree of shift progressively decreasing to zero at approximately step 100.}}

    Inspired by this phenomenon, during the progressive construction of the confidence distribution in the training phase, we primarily use the current step's distribution to perform shift correction on the global confidence, which is detailed in \autoref{sec:method_distribution_construct}.

\subsection{Consistency Reward Hacking During TTT Caused by Voting Strategy}
\label{sec:preliminary_reward_hacking}
    Most existing work on improving reasoning capabilities during the test phase primarily uses consensus-based voting mechanisms to address the lack of reward signals. However, applying scaling strategies based on fixed-parameter models to the training phase with dynamically updated parameters naturally introduces consistency reward hacking from voting strategies. Although TTRL~\citep{TTRL} refers to this phenomenon as \textit{Lucky Hit} in their paper and considers it evidence of effective negative supervision signals, we argue that this is a major cause of rapid model convergence and limited training effectiveness. Specifically, \textbf{\textit{the model tends to output consistent answers (whether correct or not) to obtain higher reward scores, which constrains the role of correctness reward.}}

    \begin{figure}[ht]
        \begin{center}
        \begin{subfigure}[b]{0.49\textwidth}
                \centering
                \includegraphics[width=\textwidth]{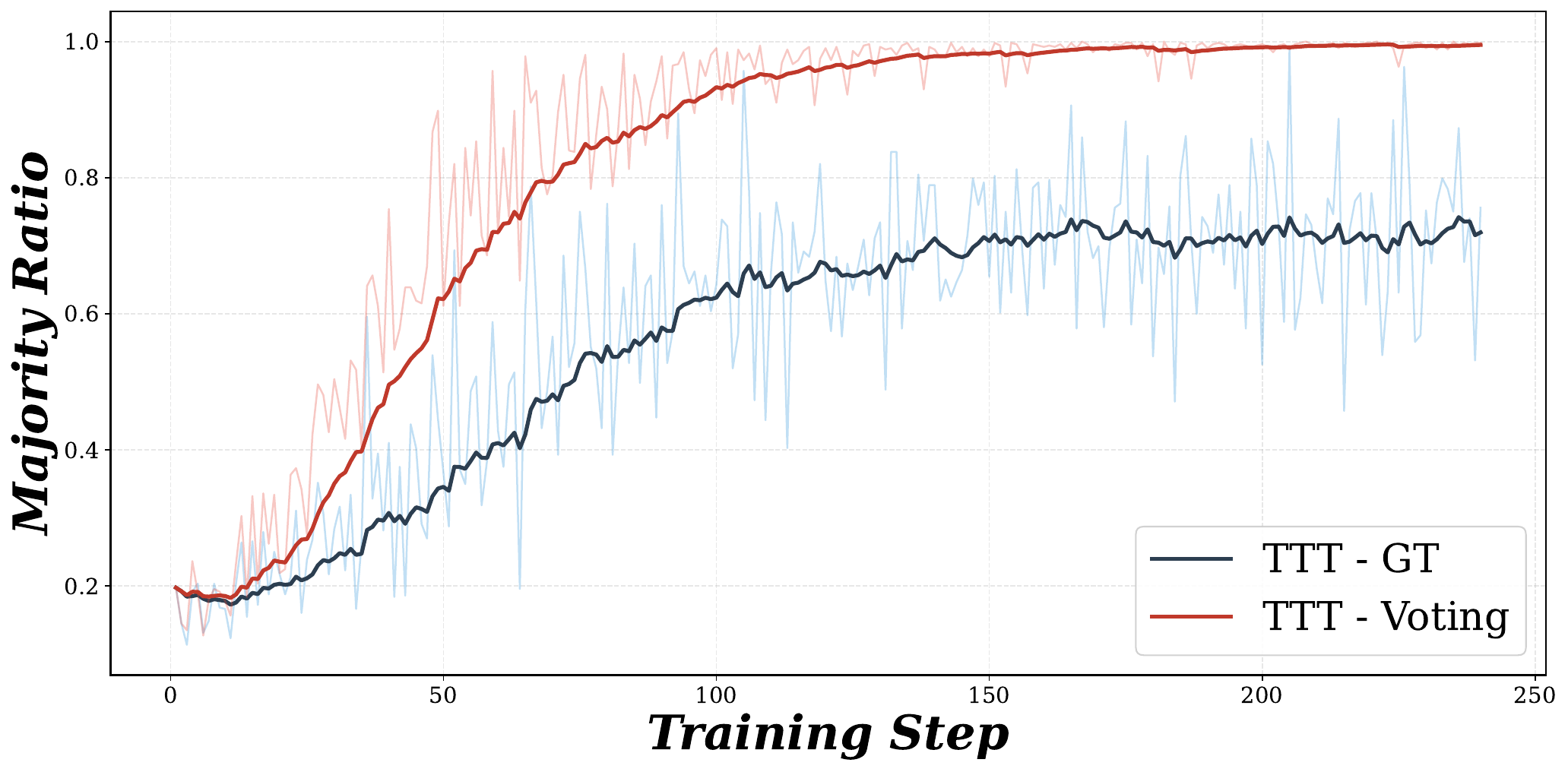}
        \end{subfigure}
        \begin{subfigure}[b]{0.49\textwidth}
                \centering
                \includegraphics[width=\textwidth]{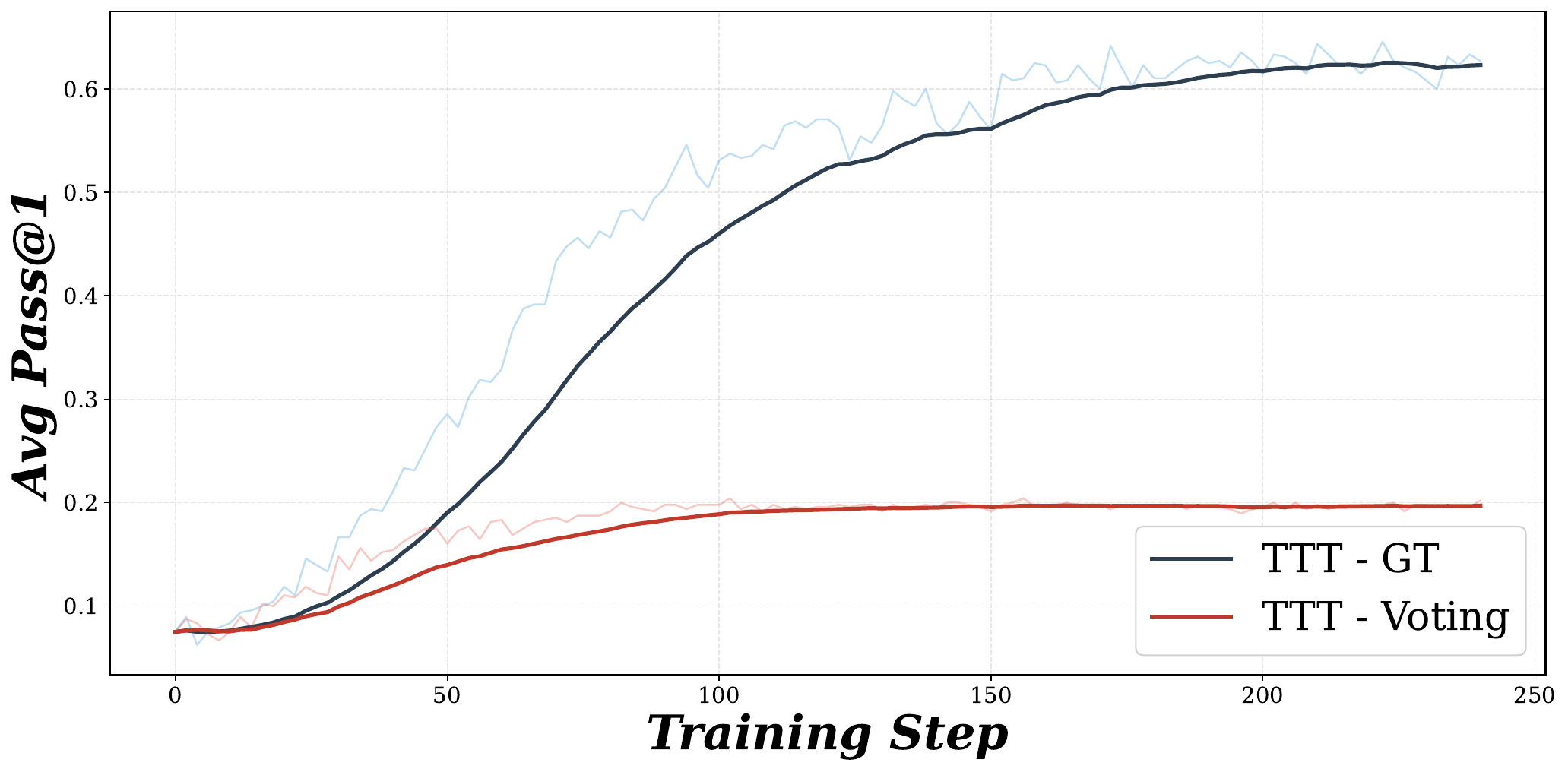}
        \end{subfigure}
        \vskip -0.1in
        \caption{
            Trend of majority ratio (left) and accuracy (right, average of 16) during the TTT process of training Qwen2.5-7B on AIME2024. GT denotes direct supervision with ground truth, while Voting denotes using Majority Voting for pseudo-label estimation.
        }
        \vskip -0.1in
        \label{fig:reward_hacking}
        \end{center}
    \end{figure}

    To demonstrate the existence and impact of this phenomenon, we track the majority ratio during training, defined as the frequency of the selected pseudo-label among all rollouts:
    \begin{equation}
    \label{eq:majority_ratio}
        r_{\text{maj}} = \frac{n_{\text{maj}}}{N}
    \end{equation}
    where $n_{\text{maj}}$ is the count of the most frequent answer and $N$ is the total number of rollouts per question.

    As shown in the \autoref{fig:reward_hacking}, the majority ratio for pseudo-label training increases significantly faster than that of ground truth supervision as training progresses. However, this enhanced consistency does not translate into improved accuracy. Instead, after the ratio rapidly reaches its upper limit, the accuracy also begins to converge (around step 125). This phenomenon indicates that the consistency voting-based TTT approach, when estimating pseudo-labels, favors consistent incorrect answers rather than remaining faithful to correct answers.

\section{Methodology}

\subsection{Confidence Distribution Modeling}
    Based on \autoref{sec:preliminary_conf_distribution}, we model the unlabeled rollouts using GMM, approximating their bimodal distribution (positive and negative reasoning paths) with two Gaussian components:
    \begin{equation}
        p(x) = \pi_1 \mathcal{N}(x|\mu_1, \sigma_1^2) + \pi_2 \mathcal{N}(x|\mu_2, \sigma_2^2),
    \end{equation}

    where $\pi_1, \pi_2$ are mixing weights satisfying $\pi_1 + \pi_2 = 1$ and $\pi_1, \pi_2 > 0$, and each component $\mathcal{N}(x|\mu_i, \sigma_i^2)$ for $i \in \{1,2\}$ follows a normal distribution:
    \begin{equation}
        \mathcal{N}(x|\mu_i, \sigma_i^2) = \frac{1}{\sqrt{2\pi\sigma_i^2}} \exp\left(-\frac{(x-\mu_i)^2}{2\sigma_i^2}\right).
    \end{equation}
    
    The mixture distribution form as follow, where $\mu_1, \mu_2$ and $\sigma_1^2, \sigma_2^2$ are the means and variances of the two distributions:
    \begin{equation}
        \label{eq:gmm}
        p(x) = \pi_1 \frac{1}{\sqrt{2\pi\sigma_1^2}} \exp\left(-\frac{(x-\mu_1)^2}{2\sigma_1^2}\right) + \pi_2 \frac{1}{\sqrt{2\pi\sigma_2^2}} \exp\left(-\frac{(x-\mu_2)^2}{2\sigma_2^2}\right).
    \end{equation}

\subsection{Progressively Pseudo-Label Estimation}
    \subsubsection{Distribution Construction}
    \label{sec:method_distribution_construct}
    During the training process, we maintain a global variable $\mathcal{C} \in \mathbb{R}^{S\times B \times G}$ to record the confidence of each rollout, where $S$, $B$, and $G$ denote the number of training steps, the number of queries in each batch, and the number of samples per query, respectively. Note that, for simplicity, we assume that each step uses only one batch for gradient update.

    For pseudo-label estimation of a specific query $q_i$ at the current step $k \in [1,\dots,S]$, we first model the distribution using the confidences $\mathcal{C}_{k,\cdot,\cdot} \in \mathbb{R}^{B\times G}$ of all queries at step $k$
    according to the GMM in \autoref{eq:gmm}. We then partition the samples into positive and negative subsets based on the fitted Gaussian components:
    \begin{equation}
    \label{eq:gmm_split}
        X_{\text{pos}}^{k} \sim \mathcal{N}\left(\mu_{\text{pos}}^{k}, {(\sigma_{\text{pos}}^{k})}^2\right), \quad
        X_{\text{neg}}^{k} \sim \mathcal{N}\left(\mu_{\text{neg}}^{k}, {(\sigma_{\text{neg}}^{k})}^2\right),
    \end{equation}
    where $\mu_{\text{pos}}^{k} = \max(\mu_1^{k}, \mu_2^{k})$ and $\mu_{\text{neg}}^{k} = \min(\mu_1^{k}, \mu_2^{k})$ correspond to the means of the high-confidence (positive) and low-confidence (negative) components, respectively.

    To leverage historical rollouts while accounting for distribution shift caused by policy updates, we apply a shift correction to the confidence values from previous steps. Specifically, for each previous step $s < k$, we fit GMMs to all queries' confidences $\mathcal{C}_{s,\cdot,\cdot}$ to get more reliable distribution, and then obtain the corresponding positive and negative component means $\mu_{\text{pos}}^{s}$ and $\mu_{\text{neg}}^{s}$. We then compute the shift offsets:
    \begin{equation}
    \label{eq:shift_correction}
        \Delta_{s \to k} = \frac{\mu_{\text{pos}}^{k} + \mu_{\text{neg}}^{k}}{2} - \frac{\mu_{\text{pos}}^{s} + \mu_{\text{neg}}^{s}}{2}.
    \end{equation}
    
    The corrected confidence from step $s$ are then:
    \begin{equation}
    \label{eq:corrected_conf}
        \tilde{\mathcal{C}}_{s,\cdot,\cdot} = \mathcal{C}_{s,\cdot,\cdot} + \Delta_{s \to k}
    \end{equation}

    Finally, for each query $q_i$ at step $k$, we construct the aggregated confidence distribution by combining the current step's confidences with the shift-corrected historical confidences:
    \begin{equation}
    \label{eq:aggregated_conf}
        \mathcal{C}_{\text{agg}}^{k} = \left\{\mathcal{C}_{k,\cdot,\cdot}\right\} \cup \left\{\tilde{\mathcal{C}}_{s,\cdot,\cdot} : s \in [1, k-1]\right\},
    \end{equation}
    which provides a more stable and data-rich prior for pseudo-label estimation at step $k$. This aggregated confidences $\mathcal{C}_{\text{agg}}^{k}$ is then used for the voting process.

    \subsubsection{Pseudo-Label Estimation}
    After obtaining the aggregated confidence distribution $\mathcal{C}_{\text{agg}}^{k}$ at step $k$, we refer to DistriVoting~\citep{DistriVoting} and repeat the operations of \autoref{eq:gmm} and \autoref{eq:gmm_split}, but replace $\mathcal{C}_{k,\cdot,\cdot}$ with $\mathcal{C}_{\text{agg}}^{k}$, thereby obtaining the global positive and negative distributions:
    \begin{equation}
    \label{eq:global_gmm_split}
        \tilde{X}_{\text{pos}}^{k} \sim \mathcal{N}\left(\tilde{\mu}_{\text{pos}}^{k}, {(\tilde{\sigma}_{\text{pos}}^{k})}^2\right), \quad
        \tilde{X}_{\text{neg}}^{k} \sim \mathcal{N}\left(\tilde{\mu}_{\text{neg}}^{k}, {(\tilde{\sigma}_{\text{neg}}^{k})}^2\right).
    \end{equation}

    For each query $q_i$ at step $k$, we then partition its samples into positive and negative subsets based on these global distributions. Specifically, for each sample $j \in [1, G]$ of query $q_i$, we assign it to the positive or negative subset according to:
    \begin{equation}
    \label{eq:sample_assignment}
        \mathcal{C}_{k,i,j} \in \begin{cases}
            X_{\text{pos}}^{k,i}, & \text{if } p(\mathcal{C}_{k,i,j} \mid \tilde{X}_{\text{pos}}^{k}) > p(\mathcal{C}_{k,i,j} \mid \tilde{X}_{\text{neg}}^{k}) \\
            X_{\text{neg}}^{k,i}, & \text{otherwise}
        \end{cases}
    \end{equation}
    where $p(\mathcal{C}_{k,i,j} \mid \tilde{X}_{\text{pos}}^{k})$ and $p(\mathcal{C}_{k,i,j} \mid \tilde{X}_{\text{neg}}^{k})$ denote the likelihood of confidence $\mathcal{C}_{k,i,j}$ under the positive and negative Gaussian distributions, respectively. This yields query-specific positive and negative sample sets $X_{\text{pos}}^{k,i}$ and $X_{\text{neg}}^{k,i}$ for pseudo-label assignment.

    Following DistriVoting, we treat $X_{\text{pos}}^{k,i}$ as high-potential correct samples and $X_{\text{neg}}^{k,i}$ as low-potential samples to be filtered out. To further mitigate the overlap between the two distributions, we first vote on the negative subset with their corresponding trajectories $\mathcal{V}_{\text{neg}}^{k,i} = \{o_{k,i,j} \mid \mathcal{C}_{k,i,j} \in X_{\text{neg}}^{k,i}\}$ to identify the most likely incorrect answer:
    \begin{equation}
    \label{eq:neg_vote}
        A_{\text{neg}}^{k,i} = f_{\text{vote}}(\mathcal{V}_{\text{neg}}^{k,i}, -X_{\text{neg}}^{k,i}),
    \end{equation}
    where $f_{\text{vote}}$ can be any voting methods, we simply use the majority voting here. We then remove false positive samples from $X_{\text{pos}}^{k,i}$ whose trajectories produce $A_{\text{neg}}^{k,i}$:
    \begin{equation}
    \label{eq:reject_filter}
        \hat{X}_{\text{pos}}^{k,i} = \{c \mid c \in X_{\text{pos}}^{k,i}, \text{answer}(o_c) \neq A_{\text{neg}}^{k,i}\},
    \end{equation}
    and obtain the filtered trajectories $\hat{\mathcal{V}}_{\text{pos}}^{k,i} = \{o_{k,i,j} \mid \mathcal{C}_{k,i,j} \in \hat{X}_{\text{pos}}^{k,i}\}$. Finally, we vote on the filtered positive subset to obtain the pseudo-label:
    \begin{equation}
    \label{eq:final_vote}
        A_{\text{final}}^{k,i} = f_{\text{vote}}(\hat{\mathcal{V}}_{\text{pos}}^{k,i}, \hat{X}_{\text{pos}}^{k,i}).
    \end{equation}

    Trajectories producing $A_{\text{final}}^{k,i}$ are assigned positive labels for policy optimization.

\subsection{Diversity-Targeted Penalties in GRPO}

    \subsubsection{Diversity Estimation}
    Given a training batch $\mathcal{B} = \{q_1, q_2, \dots, q_B\}$ of prompts, we sample $G$ candidate outputs for each query $q_i$ from the old policy:
    \begin{equation}
        o_{i,j} \sim \pi_{\theta_{\text{old}}}(\cdot \mid q_i), \quad j=1,\dots,G.
    \end{equation}
    
    We measure the \emph{answer diversity} for each query as the number of unique outputs:
    \begin{equation}
        D(q_i) = \left|\left\{o_{i,j} : j=1,\dots,G\right\}_{\text{unique}}\right|,
    \end{equation}
    where $|\cdot|_{\text{unique}}$ denotes the cardinality of the set of distinct elements. Higher $D(q_i)$ indicates greater model diversity about query $q_i$.
    
    To obtain batch-normalized uncertainty weights, we apply a softmax transformation over the diversity scores:
    \begin{equation}
        \tilde{\mathcal{D}}(q_i) = \frac{\exp(D(q_i))}{\sum_{k=1}^{B} \exp(D(q_k))}.
    \end{equation}
    
    We then introduce a penalty mechanism to down-weight queries with extremely low diversity, which may indicate premature convergence or mode collapse. Specifically, we define the final diversity weight as:
    \begin{equation}
        \mathcal{D}(q_i) = \begin{cases}
            \tilde{\mathcal{D}}(q_i), & \text{if } D(q_i) \leq \tau \cdot G \\
            1, & \text{otherwise}
        \end{cases}
        \label{eq:uncertainty-weight}
    \end{equation}
    where $\tau \in (0,1)$ is a threshold hyperparameter (we use $\tau=0.1$ in our experiments). Queries with diversity above the threshold receive a neutral weight of $1$, preserving the original GRPO learning dynamics, while low-diversity queries receive reduced weights proportional to their relative diversity within the batch.

    \subsubsection{Diversity-Targeted Advantage Calculation}
    For each sampled output $o_{i,j}$, we obtain a reward score $R_{i,j}$ from the reward function. Following GRPO, we compute the normalized advantage within each query's sample group:
    \begin{equation}
        \hat{A}_{i,j} = \frac{R_{i,j} - \frac{1}{G}\sum_{k=1}^{G} R_{i,j}}{\sqrt{\frac{1}{G}\sum_{k=1}^{G}\left(R_{i,j} - \bar{R}_i\right)^2}},
    \end{equation}
    where $\bar{R}_i = \frac{1}{G}\sum_{k=1}^{G} R_{i,k}$. When the standard deviation is zero, we set $\hat{A}_{i,j} = 0$.
    
    We then apply the diversity weighting to obtain the final advantage:
    \begin{equation}
        \hat{A}_{i,j}' = \hat{A}_{i,j} \cdot \mathcal{D}(q_i).
        \label{eq:weighted-advantage}
    \end{equation}
    
    Note that all tokens $t$ in output $o_{i,j}$ share the same weighted advantage: $\hat{A}_{i,j,t}' = \hat{A}_{i,j}'$.

    \subsubsection{The GRPO Objective}
    Let $r_{i,j,t}(\theta)$ denote the probability ratio at token position $t$:
    \begin{equation}
        r_{i,j,t}(\theta) = \frac{\pi_{\theta}(o_{i,j,t} \mid q_i, o_{i,j,<t})}{\pi_{\theta_{\text{old}}}(o_{i,j,t} \mid q_i, o_{i,j,<t})}.
    \end{equation}
    
    The GRPO objective is then defined as:
    \begin{small}
    \begin{align}
        \mathcal{J}_{\text{GRPO}}(\theta) 
        = \mathbb{E}_{q_i \sim P(Q)} \mathbb{E}_{\{o_{i,j}\}_{j=1}^{G} \sim \pi_{\theta_{\text{old}}}} &\Bigg[
        \frac{1}{G} \sum_{j=1}^{G} \frac{1}{|o_{i,j}|} \sum_{t=1}^{|o_{i,j}|} \Bigg(
        \min\Big(
        r_{i,j,t}(\theta) \cdot \hat{A}_{i,j}\cdot \textcolor{red}{\mathcal{D}(q_i)}, \nonumber \\
        & \text{clip}\big(r_{i,j,t}(\theta), 1-\epsilon, 1+\epsilon\big) \cdot \hat{A}_{i,j} \cdot \textcolor{red}{\mathcal{D}(q_i)}
        \Big)
        - \beta \cdot \widehat{D}_{\mathrm{KL}}^{(i,j,t)}
        \Bigg)
        \Bigg],
        \label{eq:igrpo-objective}
    \end{align}
    \end{small}where $\epsilon$ is the PPO clipping parameter, $\beta$ is the KL regularization coefficient, and $\widehat{D}_{\mathrm{KL}}^{(i,j,t)}$ is the per-token KL divergence estimator.
\section{Experiments}

    \subsection{Implementation Details}
        Regarding the implementation details of our methods, we primarily follow the codebase of TTRL, using AIME2024, AMC and MATH-500 as training and test sets, with GRPO as the base RL algorithm. For the temperature in the sampling process, we set it to 1.0 for Qwen2.5-Math and LRMs such as Qwen3-8B, and 0.6 for all others. For max response length, we set it to 32K for LRMs and 3K for other non-reasoning models. Additionally, we continue the rollout downsampling operation from TTRL, where we sample 64 trajectories and select 32 for gradient updates. All experiments were conducted on NVIDIA H-Series GPU. Complete pseudo code is provided in \autoref{sec:appendix_pseudo_code}.
        
    \subsection{Main Results}
        \begin{table}[t]
    \caption{Main Results of \textit{DistriTTRL} across benchmarks (16 repeats). -GT denotes using ground truth directly as labels to measure the upper bound of the method. -WSC denotes using Weighted Self-Consistency (Weighted Majority Voting) as the voting method, while -GMM denotes using \textit{DistriVoting} as the voting method. \textbf{Bold} and \uline{underline} respectively indicate the optimal and suboptimal results excluding GT.}
    \label{tab:main_results}
    \begin{center}
    \resizebox{0.95\textwidth}{!}{
        \begin{tabular}{llcccc}
            \toprule
            Model  & Method & AIME2024 & AMC & MATH-500 & Avg.\\
            \midrule
            \multirow{7}{*}{Qwen2.5-7B-Base}
            & Base Model & 7.50	 & 33.06 & 59.11 & 33.22 \\
            & TTRL       & 22.08 & 52.26 & 80.38 & 51.57 \\
            & TTRL-GT    & 62.71 & 82.61 & 90.75 & 78.69 \\
            \cmidrule(lr){2-6}
            & \cellcolor{lightblue!30}TTRL-WSC   & \cellcolor{lightblue!30}\uline{23.07{\scriptsize$+$0.99}}  & \cellcolor{lightblue!30}54.07{\scriptsize$+$1.81} & \cellcolor{lightblue!30}80.30{\scriptsize$-$0.08} & \cellcolor{lightblue!30}52.48{\scriptsize$+$0.91} \\
            & \cellcolor{lightblue!60}DistriTTRL-WSC & \cellcolor{lightblue!60}22.08{\scriptsize$+$0.00} & \cellcolor{lightblue!60}\textbf{57.23{\scriptsize$+$4.97}} & \cellcolor{lightblue!60}\uline{80.67{\scriptsize$+$0.29}}  & \cellcolor{lightblue!60}\uline{53.33{\scriptsize$+$1.76}} \\
            & \cellcolor{lightblue!100}DistriTTRL-GMM & \cellcolor{lightblue!100}\textbf{23.54{\scriptsize$+$1.46}} & \cellcolor{lightblue!100}\uline{56.48{\scriptsize$+$4.22}} & \cellcolor{lightblue!100}\textbf{81.46{\scriptsize$+$1.08}} & \cellcolor{lightblue!100}\textbf{53.83{\scriptsize$+$2.26}} \\
            \midrule
            \midrule
            \multirow{7}{*}{Qwen2.5-Math-7B}
            & Base Model & 10.63 & 32.38 & 47.10 & 30.04 \\
            & TTRL       & 25.83 & 56.18 & 81.51 & 54.51 \\
            & TTRL-GT    & 41.04 & 58.43 & 84.05 & 61.17 \\
            \cmidrule(lr){2-6}
            & \cellcolor{lightblue!30}TTRL-WSC   & \cellcolor{lightblue!30}29.79{\scriptsize$+$3.96} & \cellcolor{lightblue!30}\uline{58.28{\scriptsize$+$2.10}} & \cellcolor{lightblue!30}\textbf{83.39{\scriptsize$+$1.88}} & \cellcolor{lightblue!30}57.15{\scriptsize$+$2.64} \\
            & \cellcolor{lightblue!60}DistriTTRL-WSC & \cellcolor{lightblue!60}\textbf{34.58{\scriptsize$+$8.75}} & \cellcolor{lightblue!60}57.68{\scriptsize$+$1.50} & \cellcolor{lightblue!60}81.49{\scriptsize$-$0.02} & \cellcolor{lightblue!60}\uline{57.92{\scriptsize$+$3.41}} \\
            & \cellcolor{lightblue!100}DistriTTRL-GMM & \cellcolor{lightblue!100}\uline{33.33{\scriptsize$+$7.50}} & \cellcolor{lightblue!100}\textbf{61.37{\scriptsize$+$5.19}} & \cellcolor{lightblue!100}\uline{81.83{\scriptsize$+$0.32}} & \cellcolor{lightblue!100}\textbf{58.84{\scriptsize$+$4.33}} \\
            \midrule
            \midrule
            \multirow{7}{*}{Qwen2.5-Math-1.5B}
            & Base Model & 7.08  & 28.09 & 31.41 & 22.19 \\
            & TTRL       & 14.14 & 44.38 & 72.19 & 43.57 \\
            & TTRL-GT    & 17.94 & 47.85 & 73.45 & 46.41 \\
            \cmidrule(lr){2-6}
            & \cellcolor{lightblue!30}TTRL-WSC   & \cellcolor{lightblue!30}\uline{14.28{\scriptsize$+$0.14}} & \cellcolor{lightblue!30}44.78{\scriptsize$+$0.40} & \cellcolor{lightblue!30}72.21{\scriptsize$+$0.02} & \cellcolor{lightblue!30}43.76{\scriptsize$+$0.19} \\
            & \cellcolor{lightblue!60}DistriTTRL-WSC & 
            \cellcolor{lightblue!60}14.12{\scriptsize$-$0.02} & \cellcolor{lightblue!60}\uline{44.99{\scriptsize$+$0.61}} & \cellcolor{lightblue!60}\textbf{72.49{\scriptsize$+$0.30}} & \cellcolor{lightblue!60}\uline{43.87{\scriptsize$+$0.30}} \\
            & \cellcolor{lightblue!100}DistriTTRL-GMM & \cellcolor{lightblue!100}\textbf{14.55{\scriptsize$+$0.41}} & \cellcolor{lightblue!100}\textbf{45.37{\scriptsize$+$0.99}} & \cellcolor{lightblue!100}\uline{72.31{\scriptsize$+$0.12}} & \cellcolor{lightblue!100}\textbf{44.08{\scriptsize$+$0.51}} \\
            \midrule
            \midrule
            \multirow{7}{*}{Llama-3.1-8B-Instruct}
            & Base Model & 4.79	 & 20.93 & 48.23 & 24.65 \\
            & TTRL       & 7.92	 & 30.27 & 58.43 & 32.21 \\
            & TTRL-GT    & 21.25 & 37.65 & 75.43 & 44.78 \\
            \cmidrule(lr){2-6}
            & \cellcolor{lightblue!30}TTRL-WSC   & \cellcolor{lightblue!30}\textbf{12.08{\scriptsize$+$4.16}} & \cellcolor{lightblue!30}\uline{29.59{\scriptsize$-$0.68}} & \cellcolor{lightblue!30}55.28{\scriptsize$-$3.15} & \cellcolor{lightblue!30}32.32{\scriptsize$+$0.11} \\
            & \cellcolor{lightblue!60}DistriTTRL-WSC & \cellcolor{lightblue!60}10.00{\scriptsize$+$2.08} & \cellcolor{lightblue!60}28.84{\scriptsize$-$1.43} & \cellcolor{lightblue!60}\textbf{60.93{\scriptsize$+$2.50}} & \cellcolor{lightblue!60}\uline{33.26{\scriptsize$+$1.05}} \\
            & \cellcolor{lightblue!100}DistriTTRL-GMM & \cellcolor{lightblue!100}\uline{10.21{\scriptsize$+$2.29}} & \cellcolor{lightblue!100}\textbf{30.87{\scriptsize$+$0.60}} & \cellcolor{lightblue!100}\uline{58.96{\scriptsize$+$0.53}} & \cellcolor{lightblue!100}\textbf{33.35{\scriptsize$+$1.14}} \\
            \midrule
            \midrule
            \multirow{7}{*}{Qwen3-8B}
            & Base Model & 1.46  & 15.44 & 50.13 & 22.34 \\
            & TTRL       & 33.00 & 65.95 & 87.25 & 62.07 \\
            & TTRL-GT    & 34.07 & 67.87 & 87.58 & 63.17 \\
            \cmidrule(lr){2-6}
            & \cellcolor{lightblue!30}TTRL-WSC   & \cellcolor{lightblue!30}\uline{34.55{\scriptsize$+$1.55}} & \cellcolor{lightblue!30}\uline{68.80{\scriptsize$+$2.85}} & \cellcolor{lightblue!30}86.13{\scriptsize$-$1.12} & \cellcolor{lightblue!30}63.16{\scriptsize$+$1.09} \\
            & \cellcolor{lightblue!60}DistriTTRL-WSC & \cellcolor{lightblue!60}\textbf{37.61{\scriptsize$+$4.61}} & \cellcolor{lightblue!60}66.57{\scriptsize$+$0.62} & \cellcolor{lightblue!60}\uline{86.18{\scriptsize$-$1.07}} & \cellcolor{lightblue!60}\uline{63.45{\scriptsize$+$1.38}} \\
            & \cellcolor{lightblue!100}DistriTTRL-GMM & \cellcolor{lightblue!100}33.44{\scriptsize$+$0.44} & \cellcolor{lightblue!100}\textbf{70.31{\scriptsize$+$4.36}} & \cellcolor{lightblue!100}\textbf{87.21{\scriptsize$-$0.04}} & \cellcolor{lightblue!100}\textbf{63.65{\scriptsize$+$1.58}} \\
            \bottomrule
        \end{tabular}
    }
    \end{center}
\end{table}
        In the experimental section, we evaluate \textit{DistriTTRL}'s performance across different benchmarks and models, analyzing: (1) the direct impact of training-time confidence on TTT (TTRL vs. TTRL-WSC), (2) the effect of diversity-targeted penalty applied to the advantage function (TTRL-WSC vs. DistriTTRL-WSC), and (3) the effectiveness of leveraging confidence distribution priors for pseudo-label estimation in \textit{DistriTTRL} (DistriTTRL-WSC vs. DistriTTRL-GMM). As shown in \autoref{tab:main_results}, our proposed DistriTTRL-GMM consistently outperforms baselines across five models (including Qwen2.5-7B-Base) and three benchmarks (including AIME2024), achieving improvements of $0.51\sim4.33$. 

        Specifically, the difference between TTRL-WSC and TTRL ($0.19\sim2.64$) validates the effectiveness of incorporating confidence scores in \textit{DistriTTRL}, demonstrating that leveraging model-internal confidence during training provides more reliable signals for policy optimization compared to uniform weighting. The gap between DistriTTRL-WSC and TTRL-WSC ($0.11\sim0.96$) reflects the direct contribution of diversity-targeted penalty, which mitigates consistency reward hacking by encouraging exploration of diverse reasoning paths rather than converging to repetitive solutions. Additionally, the performance gain of DistriTTRL-GMM over DistriTTRL-WSC ($0.2\sim0.92$) demonstrates the advantage of calibrating confidence with distribution priors, where dynamically modeling the confidence distribution and applying shift correction enables more accurate pseudo-label estimation by accounting for the evolving confidence landscape throughout training.
\section{Analysis}
    \subsection{Diversity-Targeted Penalty Mitigates Consistency Reward Hacking}
        To further analyze the effectiveness of \textit{DistriTTRL} in addressing consistency reward hacking, beyond the accuracy reported in the main results, we examine the trend of majority ratio defined in \autoref{eq:majority_ratio} during training, following the same analysis approach as in \autoref{sec:preliminary_reward_hacking}.

        \begin{figure}[ht]
            \begin{center}
            \centerline{\includegraphics[width=0.8\columnwidth]{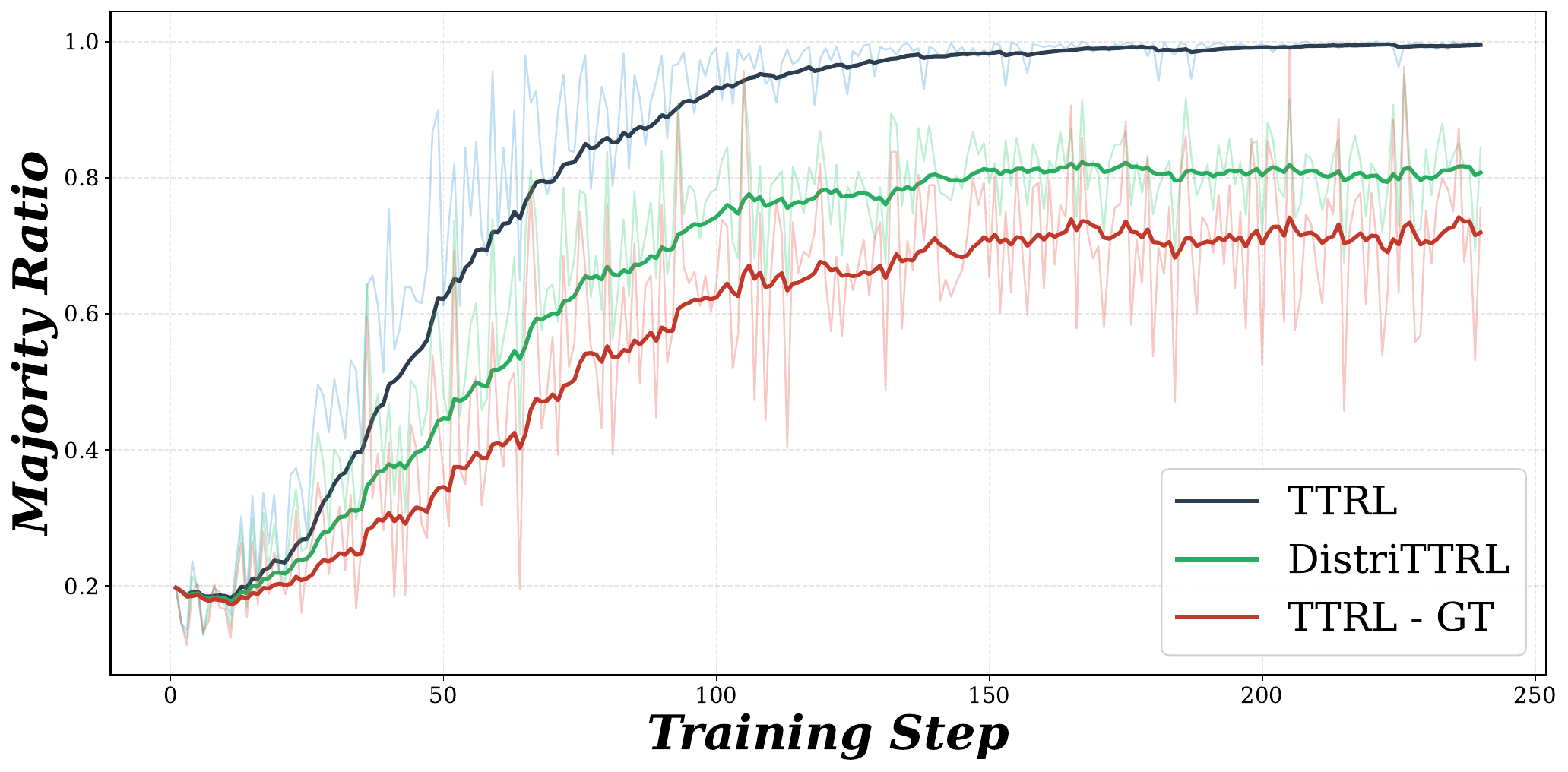}}
            \vskip -0.1in
            \caption{
                Effect of diversity-targeted penalty in \textit{DistriTTRL} on mitigating consistency reward hacking, training Qwen2.5-7B on AIME2024.
            }
            \vskip -0.2in
            \label{fig:reward_hacking_ana}
            \end{center}
        \end{figure}

        As shown in the \autoref{fig:reward_hacking_ana}, \textit{DistriTTRL}'s majority ratio curve rises more slowly and converges to a significantly lower value than TTRL (which directly uses test-time scaling strategy), better aligning with the ground truth training scenario. By mitigating consistency reward hacking, \textit{DistriTTRL} extends effective training steps and achieves higher performance gains.

    \subsection{Effect of Voting Budget on Pseudo-Label Quality}
    \label{sec:analysis_budget_scaling}
        A key motivation for \textit{DistriTTRL} stems from the observation that limited computational resources during training restrict the sampling budget per query in scaling voting strategies, thereby compromising the accuracy of pseudo-labels obtained through voting. This motivates the design of a mechanism that dynamically leverages rollouts across different steps during training. To validate this motivation, in this section we examine how the accuracy of various parallel voting strategies varies under different sampling budgets.

        \begin{figure}[ht]
            \begin{center}
            \centerline{\includegraphics[width=0.8\columnwidth]{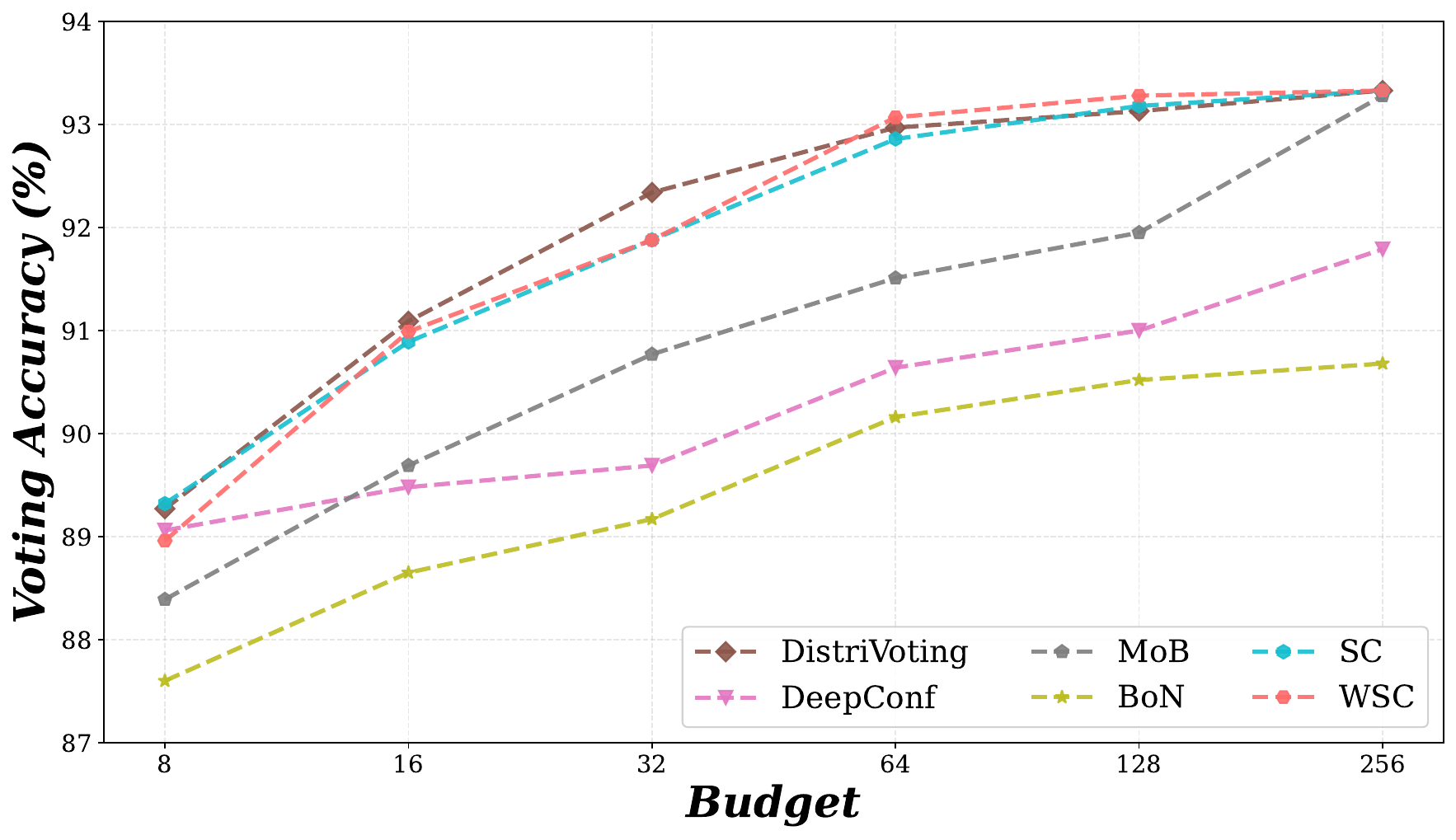}}
            \vskip -0.1in
            \caption{
                Impact of voting budget on the accuracy of different TTS strategies, evaluated on AIME2024 using DeepSeek-R1-8B (64 repeats). Complete results provided in ~\autoref{sec:appendix_budget_scaling}.
            }
            \vskip -0.2in
            \label{fig:budget_acc_ana}
            \end{center}
        \end{figure}

        As shown in the \autoref{fig:budget_acc_ana}, accuracy for all TTS strategies (SC~\citep{Self-Consistency}, WSC~\citep{WSC}, MoB~\citep{MoB}, BoN~\citep{BoN}, DistriVoting~\citep{DistriVoting}, and DeepConf~\citep{DeepConf}) increases with budget from 8 to 256, confirming a positive correlation between voting accuracy and budget. This validates \textit{DistriTTRL}'s design rationale of scaling voting information by leveraging rollouts across training steps.

\section{Conclusion}
    This paper proposes \textit{DistriTTRL}, a method that fully leverages the model's internal confidence for Test-Time Training (TTT). We first analyze the pitfalls of directly adapting scaling voting strategies for pseudo-label computation in current reasoning TTT works, namely the limited voting budget and the sensitivity of learnable parameters to consistency hacking during training. Building upon a voting strategy that calibrates confidence using distribution priors, \textit{DistriTTRL} dynamically constructs and progressively refines the confidence distribution during training, and at each step, applies shift correction to the global distribution from previous steps using the local distribution of the current step. Furthermore, leveraging the rollouts for each query, we evaluate query-level diversity and design a diversity-targeted penalty that is weighted into the original advantage. Through extensive experiments and analysis, we demonstrate the effectiveness of \textit{DistriTTRL} for reasoning TTT.

\bibliography{DistriTTRL}
\bibliographystyle{colm2026_conference}

\appendix
\newpage

\section{Pseudo Code}
\label{sec:appendix_pseudo_code}
    \begin{algorithm}
    \label{alg:framework}
    \caption{DistriTTRL Training Framework}
    \begin{algorithmic}[1]
    \footnotesize
    \Require Training queries $\mathcal{Q}$, initial policy $\pi_{\theta_0}$, reward model $R$, total steps $S$
    \Require Batch size $B$, samples per query $G$, diversity threshold $\tau$
    \State Initialize confidence storage $\mathcal{C} \in \mathbb{R}^{S \times B \times G}$
    \For{step $k = 1$ to $S$}
        \State Sample batch $\mathcal{B}_k = \{q_1, \ldots, q_B\}$ from $\mathcal{Q}$
        \For{each query $q_i \in \mathcal{B}_k$}
            \State Sample $G$ rollouts: $\{o_{k,i,j}\}_{j=1}^G \sim \pi_{\theta_{k-1}}(\cdot \mid q_i)$
            \State Compute rewards: $R_{k,i,j} = R(q_i, o_{k,i,j})$ for $j=1,\ldots,G$
            \State Compute confidences: $\mathcal{C}_{k,i,j}$ for $j=1,\ldots,G$
        \EndFor
        \State $\mathcal{C}_{\text{agg}}^k \gets$ \Call{ConstructAggregatedDistribution}{$\mathcal{C}$, $k$} \Comment{\autoref{alg:construct_distribution}}
        \For{each query $q_i \in \mathcal{B}_k$}
            \State $A_{\text{final}}^{k,i} \gets$ \Call{EstimatePseudoLabel}{$q_i$, $\mathcal{C}_{\text{agg}}^k$, $k$} \Comment{\autoref{alg:pseudo_label}}
            \State $\mathcal{D}(q_i) \gets$ \Call{ComputeDiversityWeight}{$\{o_{k,i,j}\}_{j=1}^G$, $\tau$} \Comment{\autoref{alg:diversity_weight}}
        \EndFor
        \State Update policy: $\theta_k \gets$ \Call{GRPOUpdate}{$\theta_{k-1}$, $\mathcal{B}_k$, $\{A_{\text{final}}^{k,i}\}$, $\{\mathcal{D}(q_i)\}$}
    \EndFor
    \State \Return $\pi_{\theta_S}$
    \end{algorithmic}
    \end{algorithm}

    \begin{algorithm}
    \caption{Construct Aggregated Confidence Distribution}
    \label{alg:construct_distribution}
    \begin{algorithmic}[1]
    \footnotesize
    \Function{ConstructAggregatedDistribution}{$\mathcal{C}$, $k$}
        \State Fit GMM on $\mathcal{C}_{k,\cdot,\cdot}$ via \autoref{eq:gmm_split} to obtain $\mu_{\text{pos}}^k, \mu_{\text{neg}}^k$
        \State Initialize $\mathcal{C}_{\text{agg}}^k \gets \{\mathcal{C}_{k,\cdot,\cdot}\}$
        \For{each previous step $s = 1$ to $k-1$}
            \State Fit GMM on $\mathcal{C}_{s,\cdot,\cdot}$ to obtain $\mu_{\text{pos}}^s, \mu_{\text{neg}}^s$
            \State Compute shift offset $\Delta_{s \to k}$ via \autoref{eq:shift_correction}
            \State Apply correction to get $\tilde{\mathcal{C}}_{s,\cdot,\cdot}$ via \autoref{eq:corrected_conf}
            \State $\mathcal{C}_{\text{agg}}^k \gets \mathcal{C}_{\text{agg}}^k \cup \{\tilde{\mathcal{C}}_{s,\cdot,\cdot}\}$
        \EndFor
        \State \Return $\mathcal{C}_{\text{agg}}^k$ as defined in \autoref{eq:aggregated_conf}
    \EndFunction
    \end{algorithmic}
    \end{algorithm}

    \begin{algorithm}
    \caption{Estimate Pseudo-Label via Progressive Voting}
    \label{alg:pseudo_label}
    \begin{algorithmic}[1]
    \footnotesize
    \Function{EstimatePseudoLabel}{$q_i$, $\mathcal{C}_{\text{agg}}^k$, $k$}
        \State Fit GMM on $\mathcal{C}_{\text{agg}}^k$ to obtain global distributions via \autoref{eq:global_gmm_split}
        \State Initialize $X_{\text{pos}}^{k,i} \gets \emptyset$, $X_{\text{neg}}^{k,i} \gets \emptyset$
        \For{each sample $j = 1$ to $G$}
            \State Assign $\mathcal{C}_{k,i,j}$ to $X_{\text{pos}}^{k,i}$ or $X_{\text{neg}}^{k,i}$ via \autoref{eq:sample_assignment}
        \EndFor
        \State Obtain $\mathcal{V}_{\text{neg}}^{k,i}$ and vote for $A_{\text{neg}}^{k,i}$ via \autoref{eq:neg_vote}
        \State Filter false positives to get $\hat{X}_{\text{pos}}^{k,i}$ via \autoref{eq:reject_filter}
        \State Obtain $\hat{\mathcal{V}}_{\text{pos}}^{k,i}$ and vote for $A_{\text{final}}^{k,i}$ via \autoref{eq:final_vote}
        \State \Return $A_{\text{final}}^{k,i}$
    \EndFunction
    \end{algorithmic}
    \end{algorithm}
    
    \begin{algorithm}
    \caption{Compute Diversity-Targeted Weight}
    \label{alg:diversity_weight}
    \begin{algorithmic}[1]
    \footnotesize
    \Function{ComputeDiversityWeight}{$\{o_{i,j}\}_{j=1}^G$, $\tau$}
        \State Compute answer diversity: $D(q_i) \gets |\{o_{i,j}\}_{j=1}^G|_{\text{unique}}$
        \State Compute batch-normalized weight: $\tilde{\mathcal{D}}(q_i) \gets \frac{\exp(D(q_i))}{\sum_{k=1}^B \exp(D(q_k))}$
        \If{$D(q_i) \leq \tau \cdot G$}
            \State \Return $\tilde{\mathcal{D}}(q_i)$ \Comment{Apply penalty for low diversity}
        \Else
            \State \Return $1$ \Comment{Neutral weight for sufficient diversity}
        \EndIf
    \EndFunction
    \end{algorithmic}
    \end{algorithm}

\newpage
\section{Additional Results on Voting Accuracy-Budget Correlation}
\label{sec:appendix_budget_scaling}
    In \autoref{sec:analysis_budget_scaling} of the main paper, we demonstrated the positive correlation between voting accuracy and sampling budget on AIME2024. To further validate the generalizability of this observation, we conduct the same analysis across four additional benchmarks: HMMT2025, BRUMO2025~\citep{HMMT&BRUMO}, AIME2025~\citep{AIME}, and GPQA-D~\citep{GPQA-D}. All experiments follow the same setup as in the main paper, using DeepSeek-R1-8B (DeepSeek-R1-0528-Qwen3-8B) and averaging over 64 repetitions.
    
    \begin{figure}[ht]
        \centering
        \begin{subfigure}[b]{0.48\columnwidth}
            \centering
            \includegraphics[width=\textwidth]{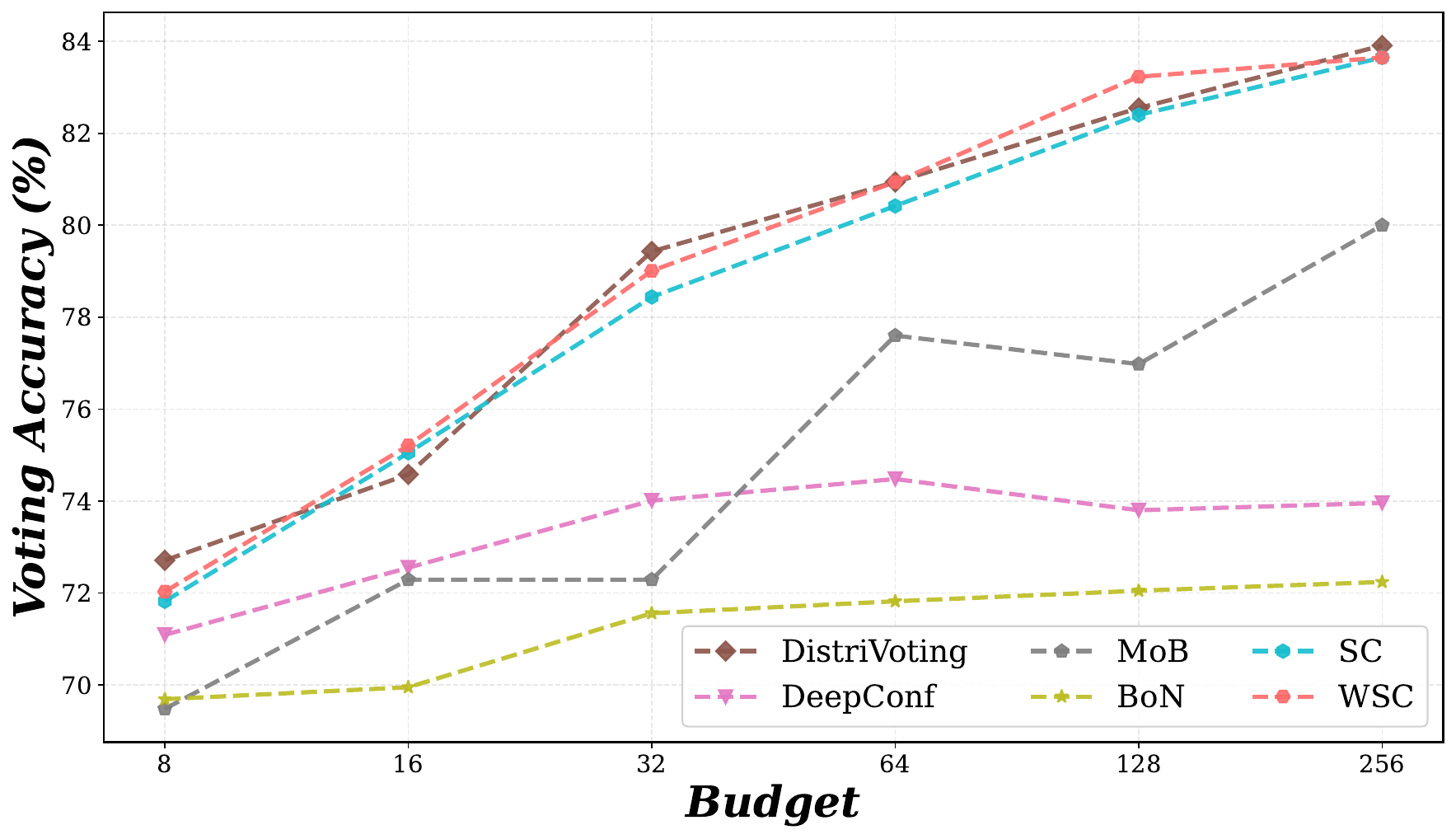}
            \caption{HMMT2025}
        \end{subfigure}
        \hfill
        \begin{subfigure}[b]{0.48\columnwidth}
            \centering
            \includegraphics[width=\textwidth]{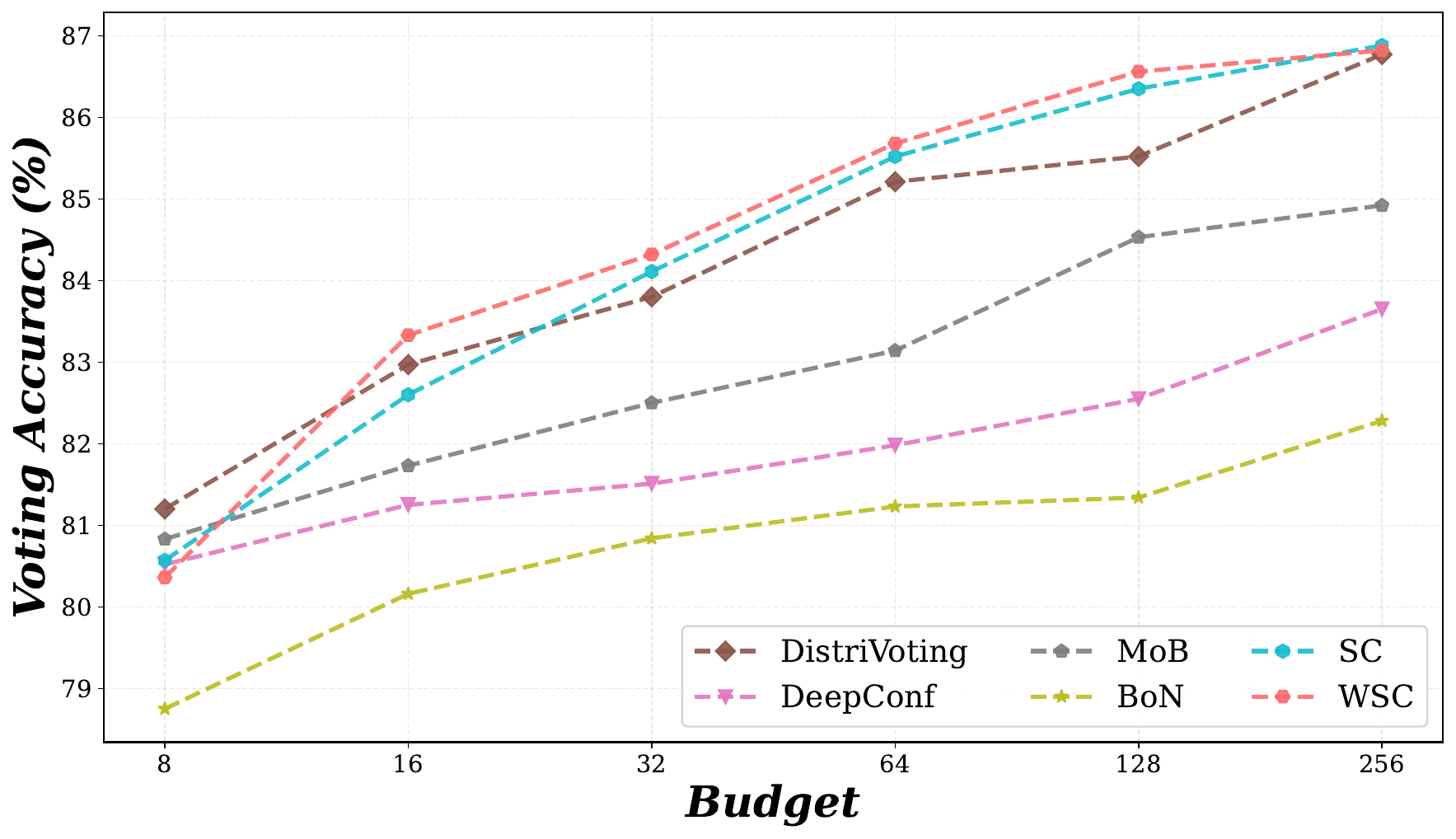}
            \caption{AIME2025}
        \end{subfigure}
        
        \vskip 0.1in
        
        \begin{subfigure}[b]{0.48\columnwidth}
            \centering
            \includegraphics[width=\textwidth]{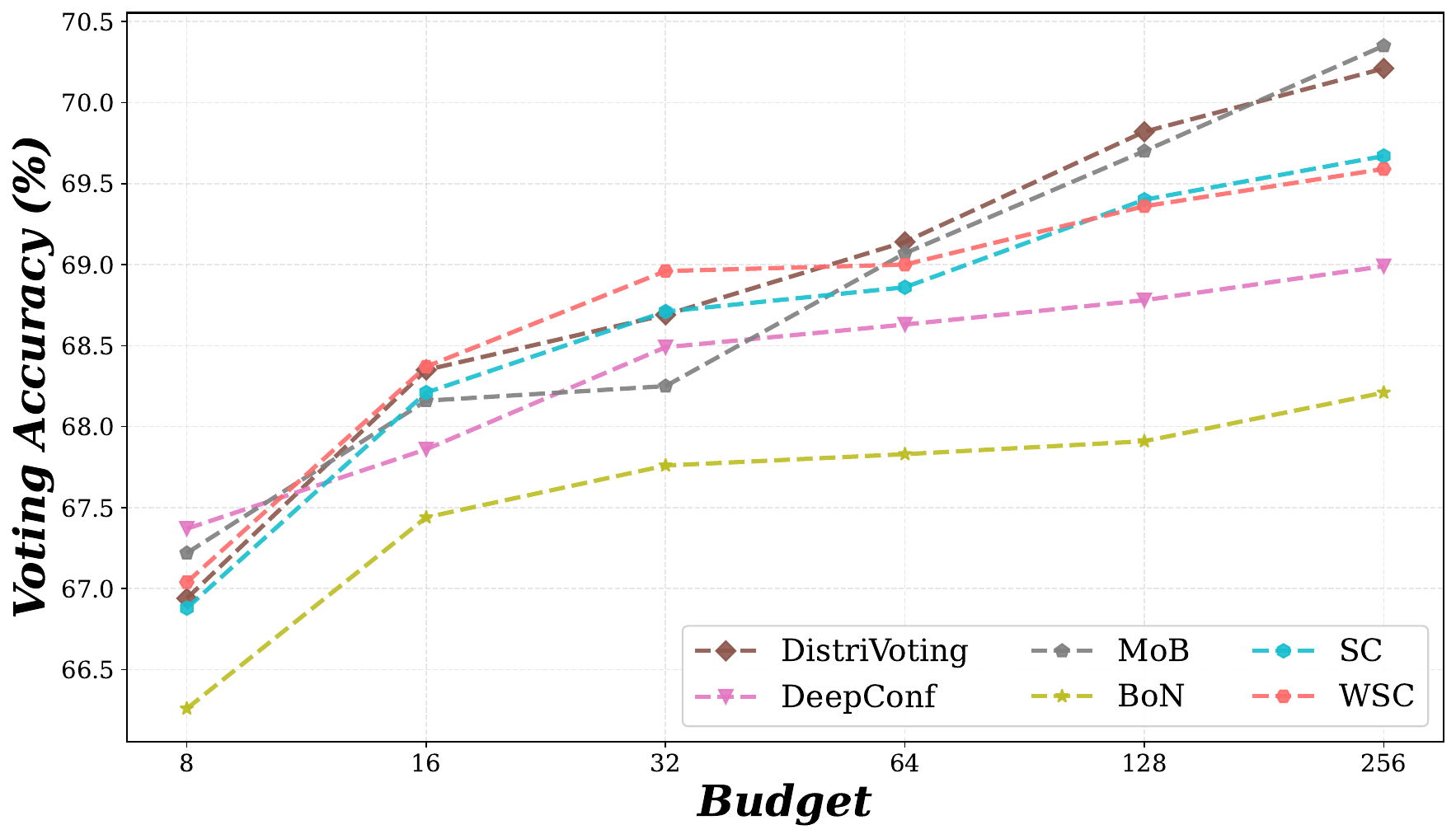}
            \caption{GPQA-D}
        \end{subfigure}
        \hfill
        \begin{subfigure}[b]{0.48\columnwidth}
            \centering
            \includegraphics[width=\textwidth]{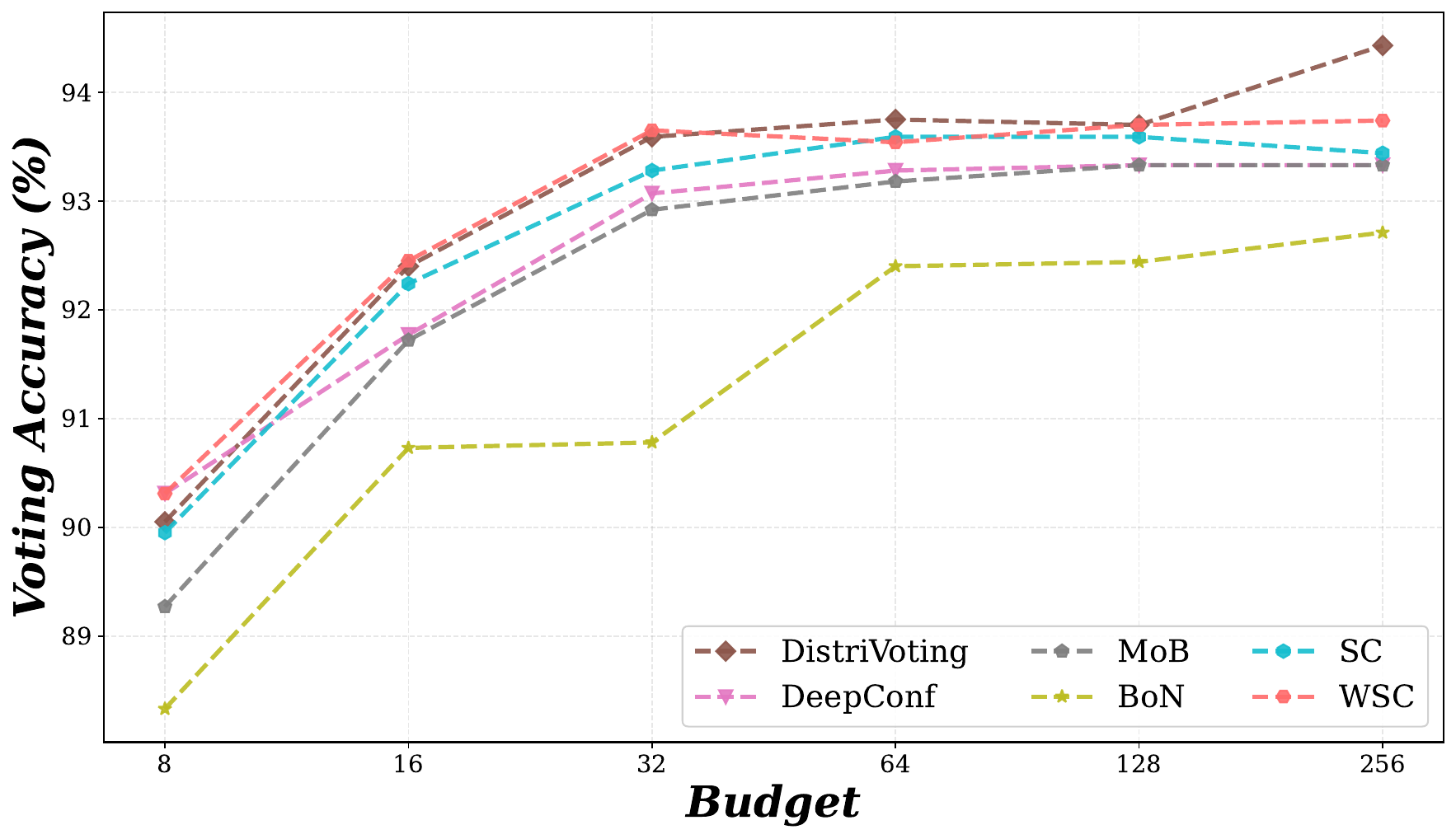}
            \caption{BRUMO2025}
        \end{subfigure}
        
        \vskip -0.1in
        \caption{
            Impact of sampling budget on the accuracy of different test-time scaling strategies across four additional benchmarks (DeepSeek-R1-8B, averaged over 64 runs).
        }
        \label{fig:budget_acc_ana_additional}
    \end{figure}

    As illustrated in \autoref{fig:budget_acc_ana_additional}, the positive correlation between voting accuracy and sampling budget consistently holds across all four benchmarks, spanning mathematical reasoning (HMMT2025, AIME2025, BRUMO2025) and scientific reasoning (GPQA-D). This validates that \textit{DistriTTRL}'s design rationale generalizes well across diverse reasoning domains. Detailed experimental results are provided in ~\autoref{tab:budget_scaling}.

    \begin{table}[ht]
    \caption{
        Complete voting accuracy comparison results of different test-time scaling strategies under varying sampling budgets across 5 benchmarks using DeepSeek-R1-8B, averaged over 64 repetitions.
    }
    \label{tab:budget_scaling}
    \begin{center}
    \resizebox{0.95\textwidth}{!}{
    \begin{tabular}{llccccc}
    \toprule
      Budget & Method & HMMT2025 & GPQA-D & AIME2024 & AIME2025 & BRUMO2025 \\
      \midrule
      \multirow{6}{*}{8}
      & MoB          & 69.48 & 67.22 & 88.39 & 80.83 & 89.27 \\
      & BoN          & 69.69 & 66.26 & 87.60 & 78.75 & 88.33 \\
      & SC           & 71.82 & 66.88 & 89.32 & 80.57 & 89.95 \\
      & WSC          & 72.03 & 67.04 & 88.96 & 80.36 & 90.31 \\
      & DeepConf     & 71.09 & 67.37 & 89.06 & 80.52 & 90.31 \\
      & DistriVoting & 72.71 & 66.94 & 89.27 & 81.20 & 90.05 \\
      \midrule
      \multirow{6}{*}{16}
      & MoB          & 72.29 & 68.16 & 89.69 & 81.73 & 91.72 \\
      & BoN          & 69.95 & 67.44 & 88.65 & 80.16 & 90.73 \\
      & SC           & 75.05 & 68.21 & 90.89 & 82.60 & 92.24 \\
      & WSC          & 75.21 & 68.37 & 90.99 & 83.33 & 92.45 \\
      & DeepConf     & 72.55 & 67.86 & 89.48 & 81.25 & 91.77 \\
      & DistriVoting & 74.58 & 68.35 & 91.09 & 82.97 & 92.40 \\
      \midrule
      \multirow{6}{*}{32}
      & MoB          & 72.29 & 68.25 & 90.77 & 82.50 & 92.92 \\
      & BoN          & 71.56 & 67.76 & 89.17 & 80.84 & 90.78 \\
      & SC           & 78.44 & 68.71 & 91.88 & 84.11 & 93.28 \\
      & WSC          & 79.01 & 68.96 & 91.88 & 84.32 & 93.65 \\
      & DeepConf     & 74.01 & 68.49 & 89.69 & 81.51 & 93.07 \\
      & DistriVoting & 79.43 & 68.69 & 92.34 & 83.80 & 93.59 \\
      \midrule
      \multirow{6}{*}{64}
      & MoB          & 77.60 & 69.07 & 91.51 & 83.14 & 93.18 \\
      & BoN          & 71.82 & 67.83 & 90.16 & 81.23 & 92.40 \\
      & SC           & 80.42 & 68.86 & 92.86 & 85.52 & 93.59 \\
      & WSC          & 80.94 & 69.00 & 93.07 & 85.68 & 93.54 \\
      & DeepConf     & 74.48 & 68.63 & 90.64 & 81.98 & 93.28 \\
      & DistriVoting & 80.94 & 69.14 & 92.97 & 85.21 & 93.75 \\
      \midrule
      \multirow{6}{*}{128}
      & MoB          & 76.98 & 69.70 & 91.95 & 84.53 & 93.33 \\
      & BoN          & 72.05 & 67.91 & 90.52 & 81.34 & 92.44 \\
      & SC           & 82.40 & 69.40 & 93.18 & 86.35 & 93.59 \\
      & WSC          & 83.23 & 69.36 & 93.28 & 86.56 & 93.70 \\
      & DeepConf     & 73.80 & 68.78 & 91.00 & 82.55 & 93.33 \\
      & DistriVoting & 82.55 & 69.82 & 93.13 & 85.52 & 93.70 \\
      \midrule
      \multirow{6}{*}{256}
      & MoB          & 80.00 & 70.35 & 93.28 & 84.92 & 93.33 \\
      & BoN          & 72.24 & 68.21 & 90.68 & 82.28 & 92.71 \\
      & SC           & 83.65 & 69.67 & 93.33 & 86.88 & 93.44 \\
      & WSC          & 83.65 & 69.59 & 93.33 & 86.82 & 93.74 \\
      & DeepConf     & 73.96 & 68.99 & 91.79 & 83.65 & 93.33 \\
      & DistriVoting & 83.91 & 70.21 & 93.33 & 86.77 & 94.43 \\
      \bottomrule
    \end{tabular}
    }
    \end{center}
\end{table}

\end{document}